\documentclass[10pt,twocolumn,letterpaper]{article}

\usepackage{cvpr}              

\usepackage[ruled,vlined,linesnumbered]{algorithm2e}

\definecolor{cvprblue}{rgb}{0.21,0.49,0.74}
\usepackage[pagebackref,breaklinks,colorlinks,allcolors=cvprblue]{hyperref}
\usepackage{comment}
\usepackage{multirow}

\def\confName{CVPR}
\def\confYear{2027}

\title{DuoCL: Dual-Path Online Continual Learning}

\author{
Jebacyril Arockiaraj$^{1}$ \qquad Dhruv Parikh$^{1}$ \qquad
Rajgopal Kannan$^{2}$ \qquad Viktor Prasanna$^{1}$\\[0.5em]
$^{1}$University of Southern California
\qquad
$^{2}$DEVCOM Army Research Office\\
{\tt\small
\{arockiar,dhruvash,prasanna\}@usc.edu
\qquad
rajgopal.kannan.civ@army.mil}
}

\begin{document}
\maketitle
\begin{abstract}
Online continual learning (OCL) requires learning from a data stream whose arrival rate is independent of model training speed. When training cannot keep pace, computationally expensive gradient-based methods update less frequently, limiting adaptation. Analytic methods reduce this overhead, but their frozen backbones cannot adapt representations to changing distributions. We propose DuoCL, a dual-path OCL framework whose fast analytic path accumulates feature statistics from every incoming minibatch while its slow path periodically adapts the backbone. 
The slow path adjusts its sample count to the available compute budget without interrupting fast-path learning.
Feature alignment compensates for drift in the accumulated classifier statistics. Across four datasets and three computational budgets, DuoCL improves online accuracy by an average of 9.67 percentage points over the strongest gradient-based baseline in each setting. DuoCL consistently outperforms analytic learning under distribution shifts, with gains of up to 4.00 percentage points on CIFAR-C, and achieves higher accuracy than prior methods across disjoint class-incremental, domain-incremental, and Si-Blurry settings.
\end{abstract}    
\section{Introduction}
\label{sec:intro}

\begin{figure*}[t]
    \centering
    \includegraphics[width=0.7\textwidth]{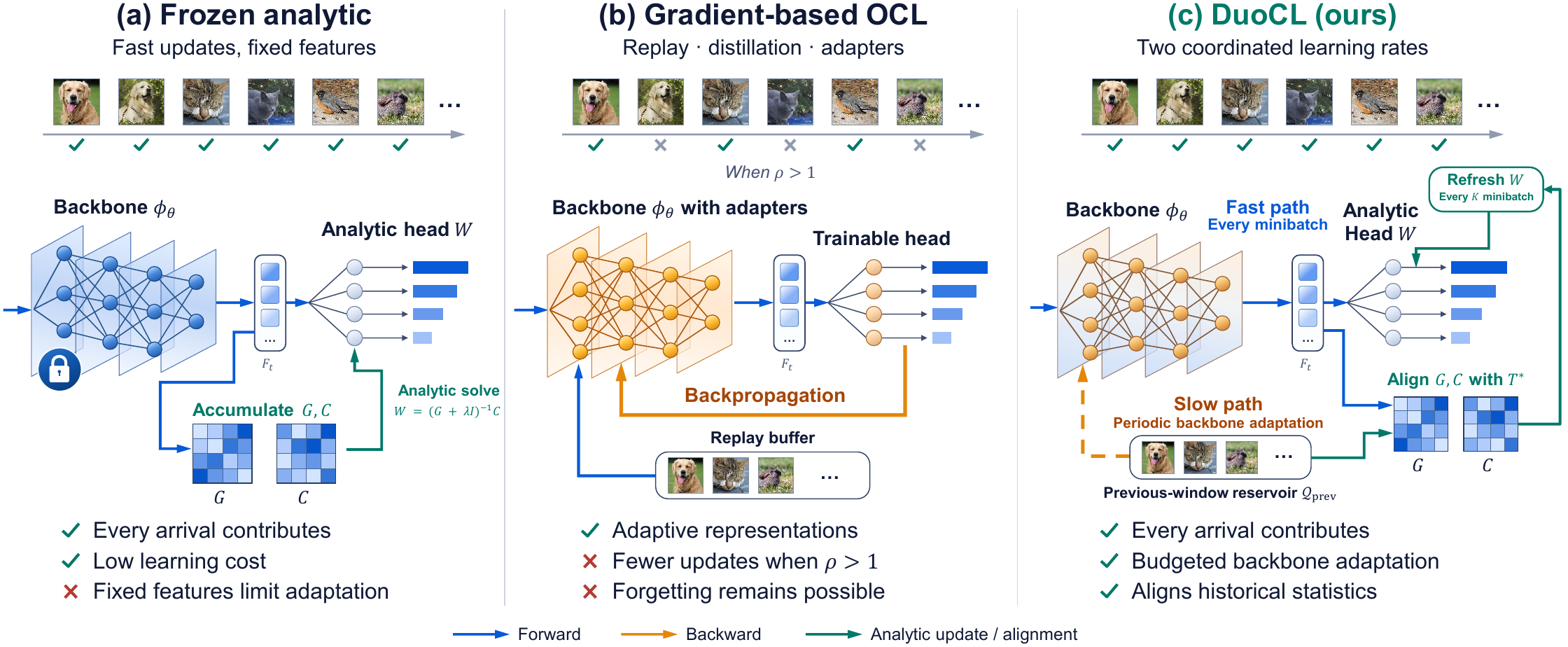}
    \caption{Comparison of OCL approaches under a computational budget. (a) Frozen analytic methods update a closed-form classifier from accumulated feature statistics $G$ and $C$ for every incoming minibatch at low cost, but the fixed backbone limits adaptation. (b) Gradient-based OCL methods adapt the backbone, but must skip minibatches for training when their computational demand exceeds the budget ($\rho > 1$). (c) DuoCL accumulates feature statistics for every incoming minibatch through the fast path, while the slow path periodically adapts the backbone using samples retained from the previous window and aligns $G$ and $C$ with the updated features.}
    \label{fig:teaser}
\end{figure*}

Continual learning (CL)~\cite{article_lifelong,de2021continual,PARISI201954} aims to enable machine learning models to learn incrementally from non-stationary data streams while mitigating catastrophic forgetting~\cite{MCCLOSKEY1989109,schwarz2018progress,de2021continual} of previously acquired knowledge. In conventional offline CL, models are trained on a sequence of tasks, typically allowing multiple training epochs over each task's data before proceeding to the next task~\cite{icarl,EWC}. In contrast, online continual learning (OCL) processes a continuous stream of data, with model updates performed under limited computational and memory budgets~\cite{CLOC}.

Existing OCL methods primarily focus on mitigating catastrophic forgetting through experience replay~\cite{Isele_Cosgun_2018, chaudhry2019continual}, regularization~\cite{10.1007/978-3-030-01219-9_9, DER, ACE}, knowledge distillation~\cite{lwf}, and sampling strategies~\cite{NEURIPS2019_15825aee, GSS, yoon2022online}. However, these methods have different computational requirements, which can limit how frequently models update when data arrives independently of model training time. A recent benchmark~\cite{Ghunaim_2023_CVPR} demonstrates that computationally expensive methods may miss learning opportunities when model training cannot keep up with the data arrival rate, with a simple baseline outperforming more sophisticated OCL methods.

Recent work~\cite{11092418} has explored approaches to support rapid adaptation under computational constraints. System-level approaches employ fine-grained pipeline parallelism to process high-frequency data streams, using gradient compensation to mitigate stale gradients. Other methods reduce training computation through selective~\cite{10350394} and adaptive layer freezing~\cite{ICLR2025_760564eb}, restricting gradient-based updates to selected backbone layers. Analytic CL methods~\cite{mcdonnell2023ranpac,loranpac2025,fed3r2024,afcl2025,stsa,arockiaraj2026} offer a different approach by freezing the backbone and updating the classifier in closed form using accumulated feature statistics. However, restricting backbone updates can limit representation plasticity, particularly when pretrained features cannot adequately capture the incoming data distribution. This motivates a central question: \textit{Can a model keep pace with the data arrival rate while periodically adapting its backbone representations under limited computational and memory budgets?}

We propose DuoCL, a dual-path OCL framework that combines continuous updates to analytic statistics with periodic gradient-based backbone adaptation (Fig.~\ref{fig:teaser}). The fast path accumulates feature statistics for every incoming minibatch, while the slow path adapts the backbone using recent samples. Coordinating these paths under limited computational and memory budgets requires determining when slow-path training occurs and how backbone updates are incorporated without interrupting fast-path learning. DuoCL addresses this through a bounded buffer of recent samples and a periodic slow-path update schedule. However, backbone adaptation changes the feature representations underlying previously accumulated classifier statistics, introducing representation drift. To mitigate this drift, DuoCL employs feature alignment to maintain consistency between the updated backbone and the analytic classifier. Our main contributions are as follows:

\begin{itemize}
    \item We propose DuoCL, a dual-path OCL framework that combines continuous analytic statistics updates with periodic gradient-based backbone adaptation.


    \item We develop a budget-aware slow path that supports periodic backbone adaptation under computational and memory constraints using bounded buffers, with feature alignment to mitigate representation drift between the updated backbone and analytic classifier statistics.

    \item Across four datasets and three computational budgets, DuoCL improves online accuracy by an average of 9.67 percentage points over the strongest gradient-based baseline in each setting.


    \item DuoCL consistently outperforms analytic learning under distribution shifts, with gains of up to 4.00 percentage points on CIFAR-C, and achieves higher accuracy than prior methods across disjoint class-incremental, domain-incremental, and Si-Blurry settings.

\end{itemize}
\section{Related Work}
\label{sec:related}

\noindent\textbf{Compute-constrained online continual learning.}
OCL methods mitigate forgetting through replay, regularization, distillation, and sample selection~\cite{chaudhry2019continual,ACE,lwf,MIR,GSS}, or adapt prompts and low-rank adapters of pretrained models online~\cite{MVP,Online-LoRA} (Fig.~\ref{fig:teaser}b). These methods typically assume that training on each minibatch completes before the next minibatch arrives. \cite{Ghunaim_2023_CVPR} evaluates OCL methods without this assumption and shows that a method $k$ times more expensive than experience replay can train on only one in every $k$ minibatches. Under such computational constraints, simple replay often outperforms more sophisticated methods~\cite{Ghunaim_2023_CVPR,prabhu2023computationally,zhang2024diet,csaba2024label}. Budget-aware methods reduce the cost of each gradient update through layer freezing~\cite{10350394,ICLR2025_760564eb} or throughput-oriented replay~\cite{wang2024forgetting}, while system-level approaches schedule retraining~\cite{bhardwaj2022ekya,khani2023recl,khani2021ams} or distribute backpropagation across additional pipeline workers~\cite{11092418}. These approaches still learn from incoming minibatches through backpropagation, so minibatches are either skipped or processed with additional computational resources. In contrast, DuoCL learns from every incoming minibatch by accumulating feature statistics using features already computed during inference, incurring little additional computation, and restricts backpropagation to a slow path whose budget is determined by the data arrival rate and device throughput (Sec.~\ref{sec:problem_defn}).

\noindent\textbf{Analytic continual learning.}
Analytic CL methods replace iterative classifier training with closed-form ridge-regression or discriminant solutions computed from accumulated feature statistics~\cite{hayes2020slda,zhuang2022acil,mcdonnell2023ranpac,loranpac2025,anacp2025}. These methods are exemplar-free and, for a fixed feature extractor, equivalent to joint training on all observed data~\cite{zhuang2022acil}. Recursive formulations update the classifier online for every minibatch~\cite{zhuang2024foal,prabhu2024randumb}, and the same statistics have been used in federated CL~\cite{fed3r2024,afcl2025,stsa,arockiaraj2026}. Such methods perform strongly under limited online budgets~\cite{prabhu2023acm,prabhu2024randumb}, but they rely on a frozen pretrained backbone, since updating it would invalidate the accumulated statistics. They therefore cannot adapt their representations when the incoming data deviate from the pretraining distribution. DuoCL addresses the limitation by adapting the backbone throughout the stream while aligning the accumulated statistics with backbone updates.

\noindent\textbf{Combining analytic and gradient-based learning.}
Fast and slow learners motivated by complementary learning systems theory~\cite{mcclelland1995cls} typically train both components with backpropagation, differing in learning rate or objective~\cite{pham2021dualnet,arani2022clser,zhang2023slca}. Other methods combine a training-free or analytic classifier with gradient-based tuning~\cite{zhao2024safe,zhang2026sfcl}, while drift compensation methods correct stored class statistics after backbone updates~\cite{yu2020sdc,goswami2024adc,magistri2024efc,rypesc2024adagauss}, including through a closed-form least-squares transformation of second-order statistics~\cite{he2025dpcr,rao2026sldc}. These methods are designed for task-based CL, where the backbone is trained for multiple epochs on each task and the statistics are corrected once per task using the full task data. SIESTA~\cite{harun2023siesta} updates its output layer online and consolidates the full network in periodic offline sleep phases, which assume scheduled downtime with no incoming data. 
To our knowledge, DuoCL is the first to combine analytic and gradient-based learning concurrently within a single-pass data stream. The fast path accumulates feature statistics for every incoming minibatch, the slow path adapts the backbone within the computational budget in every window, and a closed-form transformation estimated from a small set of retained samples aligns the accumulated classifier statistics with the backbone after each update, without requiring task boundaries, multiple training epochs, or interruptions to the stream. 

\section{Method}
\label{sec:method}

\subsection{Problem Definition}
\label{sec:problem_defn}

\begin{figure}[t]
    \centering
    \includegraphics[width=0.9\columnwidth]{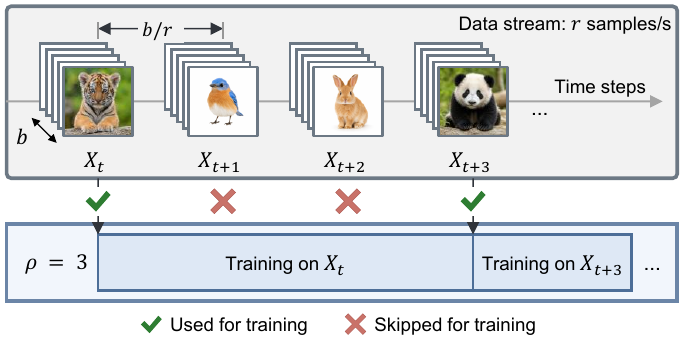}
    \caption{Problem Definition: Minibatches of $b$ samples arrive every $b/r$ seconds. With $\rho=3$, training on $X_t$ spans three arrival intervals, so $X_{t+1}$ and $X_{t+2}$ are skipped for training. Prediction continues on every arrival.}
    \label{fig:stream_setting}
\end{figure}

\noindent\textbf{Online continual learning.}
Following~\cite{CLOC,Ghunaim_2023_CVPR}, we consider a data stream arriving at $r$ samples per second and grouped into minibatches of $b$ samples. Each minibatch arrival defines a time step $t \in \{1,\ldots,T\}$, with consecutive arrivals separated by $b/r$ seconds. At step $t$, the model predicts $\hat{Y}_t = f(X_t;\theta_t)$ for the incoming minibatch $X_t$, drawn from a distribution $\pi_t$ that may change over time. The ground-truth labels $Y_t$ are then revealed to evaluate the predictions, after which the model may learn from $(X_t,Y_t)$ subject to a fixed computational budget and a memory budget of at most $M$ stored samples. Online performance is evaluated on every arrival, including minibatches skipped for training. Fig.~\ref{fig:stream_setting} illustrates the stream and an example training schedule.

\noindent\textbf{Computational budget.}
Minibatches arrive independently of training progress, so the computation that training can use between consecutive arrivals depends on both the stream rate and the device throughput. For a device providing $P$ FLOP/s for training, this computational budget per time step is
\begin{equation}
    F_{\mathrm{avail}} = \frac{Pb}{r}.
    \label{eq:available_compute}
\end{equation}
We specify $F_{\mathrm{avail}}$ directly to express different stream rates and hardware capabilities through a common computational budget. Let $F_{\mathrm{req}}$ denote the training FLOPs a method requires per time step to learn from every incoming minibatch. We characterize its demand relative to this budget by
\begin{equation}
    \rho = \frac{F_{\mathrm{req}}}{F_{\mathrm{avail}}}.
    \label{eq:compute_ratio}
\end{equation}

\noindent\textbf{Update frequency.}
When $\rho \leq 1$, a method's training procedure can keep pace with the stream. When $\rho > 1$, queuing every incoming minibatch for training would create a growing backlog that eventually exceeds the fixed memory budget. Maintaining this training procedure therefore requires skipping some minibatches. The average number of minibatches skipped per update is
\begin{equation}
    S = \max(0,\rho-1).
    \label{eq:skipped_batches}
\end{equation}
Predictions continue on every arrival using the latest available model, with parameter updates taking effect only after their computation completes.

\subsection{DuoCL Overview}
\label{sec:method_overview}

\begin{figure*}[t]
    \centering
    \includegraphics[width=0.7\textwidth]{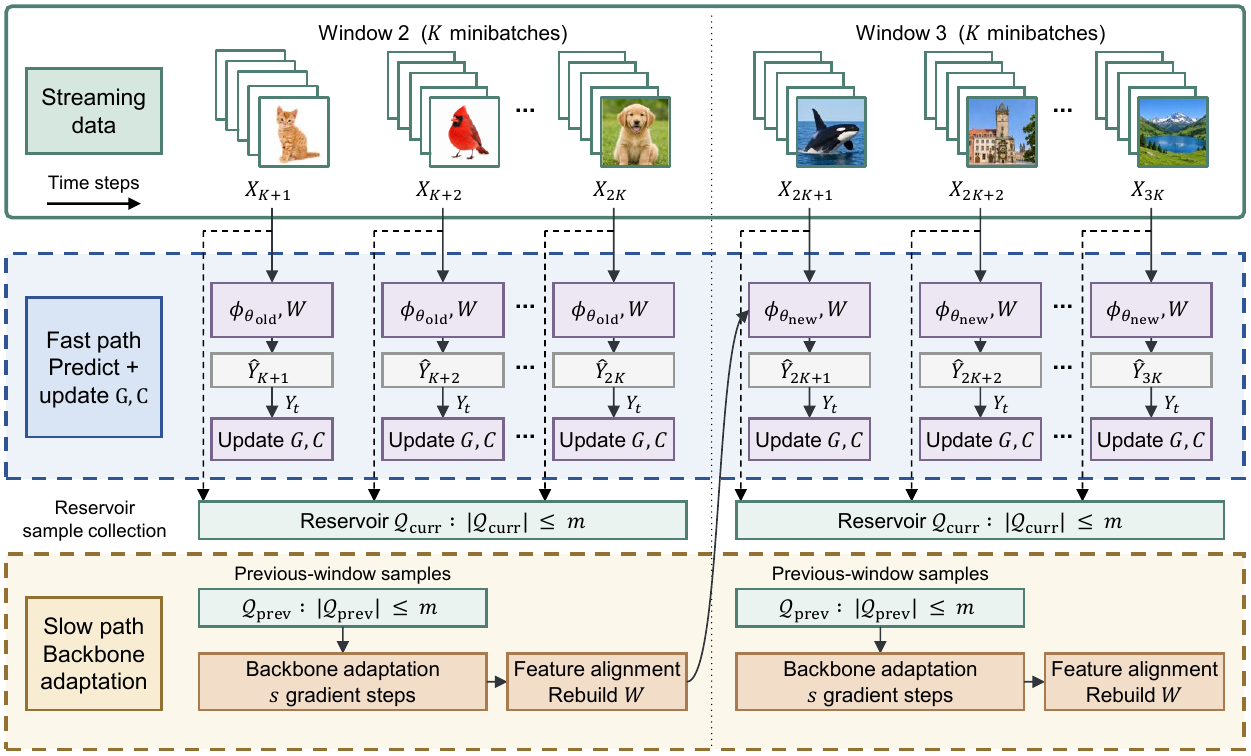}
    \caption{Overview of DuoCL over two consecutive windows of $K$ minibatches. The fast path predicts every minibatch and updates $G$ and $C$ after label revelation. Reservoir sampling collects up to $m$ current-window samples in $\mathcal{Q}_{\mathrm{curr}}$, while the slow path adapts the backbone using previous-window samples in $\mathcal{Q}_{\mathrm{prev}}$. At the next window boundary, feature alignment updates the accumulated statistics and the classifier $W$ is rebuilt.}
    \label{fig:duocl_architecture}
\end{figure*}

As illustrated in Fig.~\ref{fig:duocl_architecture}, DuoCL consists of two coordinated learning paths: a fast path for analytic classifier learning and a slow path for periodic backbone adaptation. For each incoming minibatch, the backbone $\phi_\theta$ extracts features, which the fast path incorporates into accumulated classifier statistics. The analytic classifier $W$ is periodically reconstructed from these statistics. The slow path is triggered every $K$ incoming minibatches, with each set of $K$ minibatches forming a window. During each window, the slow path adapts the backbone using samples retained from the previous window, while a new subset of samples is collected for the next update. Since backbone weight updates change the feature representations, DuoCL aligns the accumulated classifier statistics and updates the classifier. 

\subsection{Fast Path: Online Analytic Learning}
\label{sec:fast_path}

In OCL, predictions must be made for every incoming minibatch, even when the computational and memory budgets prevent training on all incoming samples. Since online prediction already requires extracting backbone features, DuoCL's fast path reuses these features and the subsequently revealed labels to accumulate feature statistics from every minibatch.

To achieve this, we adopt the ridge-regression formulation used in analytic continual learning~\cite{mcdonnell2023ranpac,loranpac2025,anacp2025}. We maintain two accumulated statistics: a Gram matrix \(G\) that captures second-order feature information, and a matrix \(C\) that accumulates class-wise feature information. For each incoming minibatch \(X_t\) containing \(b\) samples, let \(F_t=\phi_\theta(X_t)\in\mathbb{R}^{b\times d}\) denote the extracted features, where \(d\) is the feature dimension. Predictions are made using the latest available classifier $W$:
$
    \hat{Y}_t = \operatorname*{arg\,max}_j (F_t W)_{:,j},
    \label{eq:fast_prediction}
$
After observing the ground-truth labels, let \(Y_t\in\mathbb{R}^{b\times c}\) denote the one-hot encoding of the labels, where $c$ is the number of classes. We update the accumulated statistics as
$G \gets G + F_t^\top F_t$ and $C \gets C + F_t^\top Y_t$,
where $G\in\mathbb{R}^{d\times d}$ and $C\in\mathbb{R}^{d\times c}$ are initialized to zero. Each column of $C$ accumulates features belonging to a particular class, while $G$ captures second-order feature statistics used to construct the classifier. The resulting ridge-regression classifier $W\in\mathbb{R}^{d\times c}$ has the closed-form solution
\begin{equation}
    W = (G+\lambda I)^{-1}C,
    \label{eq:fast_classifier}
\end{equation}
where $\lambda>0$ is the regularization coefficient.
While $G$ and $C$ can be updated incrementally using the extracted features, reconstructing $W$ requires solving a $d\times d$ linear system, with $O(d^3)$ computational complexity for a conventional dense solve. Performing this operation after every arrival can be expensive under high data arrival rates. To reduce this cost, DuoCL updates $G$ and $C$ for every incoming minibatch but reconstructs $W$ once every $K$ minibatches. This amortizes the classifier reconstruction cost while allowing every incoming minibatch to contribute to the accumulated statistics. Unlike conventional frozen-backbone analytic learning, DuoCL also periodically adapts the backbone, requiring the accumulated statistics to be aligned with the updated representations, as in Section~\ref{sec:feature_alignment}.

\subsection{Slow Path: Periodic Backbone Adaptation}
\label{sec:slow_path}

Although the fast path incorporates every incoming minibatch into the analytic classifier statistics, it does not adapt the backbone. Performing backpropagation on every incoming minibatch can exceed the available computational budget. To address this, DuoCL periodically adapts the backbone using a bounded subset of recent samples.

\noindent\textbf{Budget-Aware Sample Selection and Scheduling.}
The slow path is triggered every $K$ incoming minibatches, with each group of $K$ minibatches forming a window. During each window, $\mathcal{Q}_{\mathrm{curr}}$ collects incoming samples, while the slow path adapts the backbone using samples retained in $\mathcal{Q}_{\mathrm{prev}}$ from the previous window. We first determine the number of samples $m$ retained in each buffer based on the available computational and memory budgets. With $b$ samples per minibatch, each window contains $Kb$ samples and provides a total computational budget of $KF_{\mathrm{avail}}$. The slow path performs $s$ gradient steps using $m$ retained samples. Let $F_{\mathrm{slow}}$ denote the profiled FLOPs required for one minibatch-equivalent of slow-path training. The retained sample count is constrained by
\begin{equation}
    \frac{sm}{b}F_{\mathrm{slow}} \leq K F_{\mathrm{avail}},
    \label{eq:slow_budget}
\end{equation}
where $KF_{\mathrm{avail}}$ denotes the available computation budget over a window. Since both buffers can coexist, their combined storage must also satisfy $2m \leq M$. Accordingly, we determine the maximum retained sample count as
\begin{equation}
    m = \min\left(
        Kb,\,
        \left\lfloor \frac{KbF_{\mathrm{avail}}}{sF_{\mathrm{slow}}} \right\rfloor,\,
        \left\lfloor \frac{M}{2} \right\rfloor
    \right).
    \label{eq:buffer_size}
\end{equation}

Since each window contains $Kb$ samples, which may exceed the buffer size $m$, we employ uniform reservoir sampling to select a subset for backbone adaptation. The first $m$ samples are stored in $\mathcal{Q}_{\mathrm{curr}}$. For each subsequent sample $n>m$, we retain it with probability $m/n$, replacing a uniformly selected sample in the buffer. This ensures that every sample in the window has an equal probability of being retained while the buffer size remains bounded by $m$. Meanwhile, the slow path performs $s$ gradient steps using samples from $\mathcal{Q}_{\mathrm{prev}}$. At the end of the window, the backbone update is applied, the previous samples are discarded, and the buffers are updated as $\mathcal{Q}_{\mathrm{prev}}\gets\mathcal{Q}_{\mathrm{curr}}$ and $\mathcal{Q}_{\mathrm{curr}}\gets\emptyset$. Thus, DuoCL performs backbone adaptation using recently retained samples without maintaining a persistent historical replay buffer.

\noindent\textbf{Slow-Path Training Objective.}
Let $(X_Q,Y_Q)$ denote the retained samples and their corresponding labels in $\mathcal{Q}_{\mathrm{prev}}$. For backbone adaptation, we employ three complementary loss terms:
\begin{equation}
    \mathcal{L}_{\mathrm{slow}} =
    \mathcal{L}_{\mathrm{main}} +
    \alpha\mathcal{L}_{\mathrm{aux}} +
    \beta\mathcal{L}_{\mathrm{drift}},
    \label{eq:slow_loss}
\end{equation}
where $\alpha$ and $\beta$ control the contributions of the auxiliary and drift losses, respectively. The main cross-entropy loss adapts the backbone features using the existing analytic classifier:
\begin{equation}
    \mathcal{L}_{\mathrm{main}} =
    \operatorname{CE}(F\widehat{W},Y_Q),
    \label{eq:main_loss}
\end{equation}
where $F=\phi_\theta(X_Q)$ denotes the features of the retained samples, $Y_Q$ denotes their labels, and $\widehat{W}$ is a fixed copy of the analytic classifier. During slow-path training, gradients propagate through the classifier to update the backbone, while $\widehat{W}$ remains unchanged.

To provide an additional supervised signal for backbone adaptation, we employ a training-only auxiliary classifier $g_\psi$, with loss
\begin{equation}
    \mathcal{L}_{\mathrm{aux}} =
    \operatorname{CE}(g_\psi(F),Y_Q).
    \label{eq:aux_loss}
\end{equation}

However, backbone updates can also change predictions for previously observed classes. To limit these changes, we employ drift regularization that penalizes deviations from the logits computed before backbone adaptation:
\begin{equation}
    \mathcal{L}_{\mathrm{drift}} =
    \left\|
    (F\widehat{W})_{:,\mathcal{O}} -
    (F_{\mathrm{pre}}\widehat{W})_{:,\mathcal{O}}
    \right\|_F^2,
    \label{eq:drift_loss}
\end{equation}
where $F_{\mathrm{pre}}$ denotes the features before adaptation and $\mathcal{O}$ denotes previously observed classes absent from the retained samples. To maintain a consistent loss scale across updates, we normalize the drift penalty by the number of retained samples and previously observed classes when applying its weight $\beta$. The drift loss is zero when $\mathcal{O}$ is empty. For CNN-based backbones, the slow path directly updates the backbone parameters. For Transformer-based backbones, we employ low-rank adaptation (LoRA)~\cite{hu2022lora}, keeping the pretrained backbone weights frozen while updating the LoRA adapters in the query and value projections. Both configurations use the same slow-path training objective.

\subsection{Drift-Aware Feature Alignment}
\label{sec:feature_alignment}

Slow-path backbone adaptation changes the feature representations used to accumulate the classifier statistics $G$ and $C$. Consequently, directly reusing these statistics with the updated backbone can introduce inconsistencies between the classifier and the new feature space. Recomputing the statistics from all previously observed samples is impractical under the computational and memory constraints of OCL. To address this, inspired by prior work on feature-space drift compensation~\cite{he2025dpcr,xu2026twoway}, DuoCL estimates a feature transformation using the retained samples and aligns the accumulated statistics with the updated backbone.

For the retained samples $X_Q$, we extract features before and after backbone adaptation:
$F_{\mathrm{pre}} = \phi_{\theta_{\mathrm{old}}}(X_Q)$ and $F_{\mathrm{post}} = \phi_{\theta_{\mathrm{new}}}(X_Q)$, where $\theta_{\mathrm{old}}$ and $\theta_{\mathrm{new}}$ denote the backbone parameters before and after adaptation, respectively. We estimate an old-to-new feature transformation $T$ using regularized least squares:
\begin{equation}
\begin{aligned}
    T^* = \arg\min_T \;&
    \left\|F_{\mathrm{pre}}T-F_{\mathrm{post}}\right\|_F^2\\
    &+\gamma\left\|T-I\right\|_F^2,
\end{aligned}
\label{eq:transport_objective}
\end{equation}
where $\gamma>0$ is the regularization coefficient that encourages the transformation to remain close to the identity matrix. This prevents excessive changes to the feature space when estimating the transformation from a limited number of retained samples. The optimization admits the closed-form solution
\begin{equation}
    T^* =
    \left(F_{\mathrm{pre}}^\top F_{\mathrm{pre}}+\gamma I\right)^{-1}
    \left(F_{\mathrm{pre}}^\top F_{\mathrm{post}}+\gamma I\right).
    \label{eq:feature_transport}
\end{equation}

Rather than recomputing the classifier statistics from previously observed samples, we transform the accumulated statistics into the updated feature space:
$G \gets T^{*\top}GT^*$ and $C \gets T^{*\top}C$.
The transformation is applied to the accumulated statistics, including minibatches processed while the slow-path update was pending. Finally, the analytic classifier $W$ is reconstructed from the aligned statistics using Eq.~\eqref{eq:fast_classifier}. This allows DuoCL to retain information accumulated from the incoming stream while periodically adapting its backbone representations.

\section{Experiment}
\label{sec:exp}


\subsection{Experimental Setup}
\label{sec:setup}

\subsubsection{Dataset and Baseline}
\label{sec:datasets}

Our evaluation covers ten widely used image classification benchmarks. CIFAR-10 and CIFAR-100~\cite{cifar100} are standard image classification benchmarks. CLEAR-10 and CLEAR-100~\cite{lin2021clear} capture the natural temporal evolution of visual concepts, while CLOC~\cite{CLOC} provides a large-scale stream with naturally occurring distribution shifts. CIFAR-C~\cite{hendrycks2019benchmarking} evaluates robustness under common image corruptions. ImageNet-R~\cite{imagenetR} and ImageNet-S~\cite{imagenetS} evaluate visual distribution shifts through artistic renditions and sketches, respectively. CUB-200~\cite{WahCUB_200_2011} provides a fine-grained classification benchmark covering 200 bird species. CORe50~\cite{pmlr-v78-lomonaco17a} evaluates continual object recognition across sessions with varying viewpoints and environments.

We compare DuoCL against representative baselines: ER~\cite{chaudhry2019continual}, ACE~\cite{ACE}, LwF~\cite{lwf}, RWalk~\cite{Rwalk}, PoLRS~\cite{CLOC}, MIR~\cite{MIR}, GSS~\cite{GSS}, and aL-SAR~\cite{ICLR2025_760564eb}. For pretrained ViT-based comparisons, we compare against AGEM~\cite{agem}, ER~\cite{chaudhry2019continual}, MIR~\cite{MIR}, GDumb~\cite{GDumb}, DER++~\cite{DER}, PCR~\cite{PCR}, LODE (with DER++~\cite{DER})~\cite{LODE}, EMA (with DER++~\cite{DER} and with RAR~\cite{RAR})~\cite{EMA}, EWC++~\cite{Rwalk}, L2P~\cite{L2P}, MVP~\cite{MVP}, Online-LoRA~\cite{Online-LoRA}, and FGH~\cite{michel2026online}. Additional implementation details, including dataset splits and hyperparameter configurations, are provided in the supplementary material.


\subsubsection{Metrics}
\label{sec:metrics}

We evaluate DuoCL using three metrics: (i) average online accuracy \(A_{\mathrm{online}}\)~\cite{Ghunaim_2023_CVPR}, the percentage of incoming samples correctly classified before learning their labels, including samples skipped for training; (ii) final accuracy \(A_{\mathrm{Final}}\)~\cite{Online-LoRA,9522763}, computed after training on the final task; and (iii) area under the curve of accuracy \(A_{\mathrm{AUC}}\)~\cite{koh2022online}, which measures model performance throughout training. We report \(A_{\mathrm{online}}\) for the online setting, \(A_{\mathrm{Final}}\) for disjoint class-incremental and domain-incremental settings, and both \(A_{\mathrm{Final}}\) and \(A_{\mathrm{AUC}}\) for Si-Blurry. See Appendix for detailed definitions.

\subsection{Results}
\label{results}

\subsubsection{Impact of Data Arrival Rate}
\label{res:online}

\begin{table}[t]
\centering
\caption{Average online accuracy (\%, $\uparrow$) under three
computational budgets, averaged over three seeds.
Improvement denotes the absolute percentage-point gain of
DuoCL over the strongest existing OCL baseline for each
dataset and budget. Best and second-best results, excluding
the unconstrained UB, are shown in \textbf{bold} and
\underline{underlined}, respectively.}
\label{tab:arrival_rate_results}
\setlength{\tabcolsep}{5.0pt}
\resizebox{\columnwidth}{!}{%
\begin{tabular}{@{}llcccc@{}}
\toprule
Method & $F_{\mathrm{avail}}$
& CIFAR-10 & CIFAR-100 & CLEAR-10 & CLEAR-100 \\
\midrule
ER~\cite{chaudhry2019continual}
& High   & \underline{64.35} & \underline{28.56} & 85.25 & 73.06 \\
& Medium & \underline{61.18} & \underline{25.04} & 80.66 & 66.89 \\
& Low    & 54.04 & \underline{19.43} & 69.09 & 52.19 \\
\midrule
ACE~\cite{ACE}
& High   & 60.99 & 24.03 & \underline{87.04} & \underline{76.63} \\
& Medium & 57.70 & 21.85 & 83.18 & \underline{72.36} \\
& Low    & 51.98 & 17.54 & 74.99 & \underline{62.66} \\
\midrule
LwF~\cite{lwf}
& High   & 63.17 &  9.90 & 82.48 & 69.58 \\
& Medium & 58.25 &  8.15 & 78.39 & 62.24 \\
& Low    & 51.31 &  6.43 & 65.71 & 44.71 \\
\midrule
RWalk~\cite{Rwalk}
& High   & 57.63 & 9.16 & 78.01 & 60.56 \\
& Medium & 54.11 & 8.19 & 69.05 & 51.90 \\
& Low    & 46.83 & 5.87 & 51.67 & 30.55 \\
\midrule
PoLRS~\cite{CLOC}
& High   & 45.47 & 17.48 & 62.72 & 48.68 \\
& Medium & 44.57 & 14.46 & 59.60 & 45.71 \\
& Low    & 41.81 & 11.11 & 18.47 & 21.23 \\
\midrule
MIR~\cite{MIR}
& High   & 49.25 & 8.18 & 72.29 & 50.53 \\
& Medium & 44.90 & 7.22 & 61.04 & 37.84 \\
& Low    & 39.02 & 4.85 & 39.05 & 17.36 \\
\midrule
GSS~\cite{GSS}
& High   & 40.26 & 9.67 & 51.43 & 30.60 \\
& Medium & 34.31 & 7.93 & 31.66 & 18.82 \\
& Low    & 28.18 & 6.09 & 16.07 &  4.76 \\
\midrule
aL-SAR~\cite{ICLR2025_760564eb}
& High   & 61.62 & 23.96 & 84.69 & 65.58 \\
& Medium & 60.17 & 22.48 & \underline{83.73} & 63.33 \\
& Low    & \underline{55.73} & 18.92 & \underline{80.57} & 58.07 \\
\midrule
DuoCL (Ours)
& High   & \textbf{64.80} & \textbf{36.64}
         & \textbf{92.70} & \textbf{83.96} \\
& Medium & \textbf{64.68} & \textbf{36.60}
         & \textbf{92.67} & \textbf{83.94} \\
& Low    & \textbf{64.26} & \textbf{36.47}
         & \textbf{92.63} & \textbf{83.92} \\
\midrule
UB
&    & 65.05 & 36.86 & 92.70 & 84.00 \\
\midrule
Improvement
& High   & +0.45 & +8.08 & +5.66 & +7.34 \\
& Medium & +3.51 & +11.56 & +8.94 & +11.59 \\
& Low    & +8.53 & +17.04 & +12.06 & +21.26 \\
\bottomrule
\end{tabular}%
}
\end{table}

To evaluate performance under varying data arrival rates, we vary the computational budget available for training per incoming minibatch, $F_{\mathrm{avail}}$, as defined in Section~\ref{sec:problem_defn}. At fixed training throughput $P$ and minibatch size $b$, a lower $F_{\mathrm{avail}}$ corresponds to a higher data arrival rate $r$. We consider three computational budgets: High, Medium, and Low, corresponding to $3\times$, $2\times$, and $1\times$ the FLOPs required for one backbone forward pass, respectively, as shown in Table~\ref{tab:arrival_budgets}. We profile the training FLOPs of each baseline and report the resulting compute ratios \(\rho\) and skipped minibatches \(S\) in Table~\ref{tab:compute_constraints}. aL-SAR~\cite{ICLR2025_760564eb} explicitly accounts for computational
cost through adaptive layer freezing. The memory budgets used in these experiments are provided in the appendix. For all methods in Table~\ref{tab:arrival_rate_results}, we use an ImageNet-1K-pretrained ResNet-18~\cite{resnet} with a batch size of 10 for CIFAR-10 and CIFAR-100, and an ImageNet-1K-pretrained ResNet-50~\cite{resnet} with a batch size of 128 for CLEAR-10 and CLEAR-100. Here, UB represents an unconstrained variant of DuoCL that retains all samples $Kb$ in each window and performs slow-path adaptation without computational restrictions.
Table~\ref{tab:arrival_rate_results} shows the average online accuracy of all methods under three computational budgets. Under the Low budget, existing OCL methods skip multiple minibatches during training, limiting their learning opportunities. In contrast, DuoCL continuously updates its accumulated feature statistics $G$ and $C$, while periodically adapting the backbone using buffered samples. As a result, DuoCL outperforms the strongest existing OCL baseline by 8.53, 17.04, 12.06, and 21.26 percentage points on CIFAR-10, CIFAR-100, CLEAR-10, and CLEAR-100, respectively, under the Low budget, with an average improvement of 9.67 percentage points across all datasets and budgets. Furthermore, DuoCL achieves accuracy within 0.79 percentage points of UB across all settings, suggesting that continuous fast-path learning combined with reservoir-based slow-path adaptation can approach unconstrained performance under limited computational and memory budgets.

\begin{table}[t]
\centering
\caption{Available computational budget $F_{\mathrm{avail}}$ (GFLOPs per arrival). FP denotes one backbone forward pass on an incoming minibatch.}
\label{tab:arrival_budgets}
\small
\setlength{\tabcolsep}{5pt}
\begin{tabular}{@{}lcc@{}}
\toprule
Budget & CIFAR-10/100 & CLEAR-10/100 \\
\midrule
High ($3\times$ FP)   & 2.22 & 3,138.92 \\
Medium ($2\times$ FP) & 1.48 & 2,092.61 \\
Low ($1\times$ FP)    & 0.74 & 1,046.31 \\
\bottomrule
\end{tabular}
\end{table}

\begin{table}[t]
\centering
\caption{Compute ratio ($\rho$) and skipped minibatches per update ($S$) under three computational budgets.}
\label{tab:compute_constraints}
\setlength{\tabcolsep}{4.0pt}
\resizebox{\columnwidth}{!}{%
\begin{tabular}{@{}lcccccc@{}}
\toprule
& \multicolumn{3}{c}{Compute Ratio ($\rho$)}
& \multicolumn{3}{c}{Skipped Minibatches ($S$)} \\
\cmidrule(lr){2-4}
\cmidrule(l){5-7}
Method & High & Medium & Low
       & High & Medium & Low \\
\midrule
ER~\cite{chaudhry2019continual}
    & 2 & 3 & 6 & 1 & 2 & 5 \\
ACE~\cite{ACE}
    & 2 & 3 & 6 & 1 & 2 & 5 \\
LwF~\cite{lwf}
    & 2.67 & 4 & 8 & 1.67 & 3 & 7 \\
RWalk~\cite{Rwalk}
    & 4 & 6 & 12 & 3 & 5 & 11 \\
PoLRS~\cite{CLOC}
    & 6 & 9 & 18 & 5 & 8 & 17 \\
MIR~\cite{MIR}
    & 5 & 7.50 & 15 & 4 & 6.50 & 14 \\
GSS~\cite{GSS}
    & 12 & 18 & 36 & 11 & 17 & 35 \\
\midrule
DuoCL(Fast Path)
    & $<0.01$ & $<0.01$ & 0.01
    & 0 & 0 & 0 \\
\bottomrule
\end{tabular}%
}
\end{table}

\subsubsection{Impact of Distribution Shift}
\label{sec:dist_shift}

\begin{figure}
    \centering
    \includegraphics[width=0.95\columnwidth]{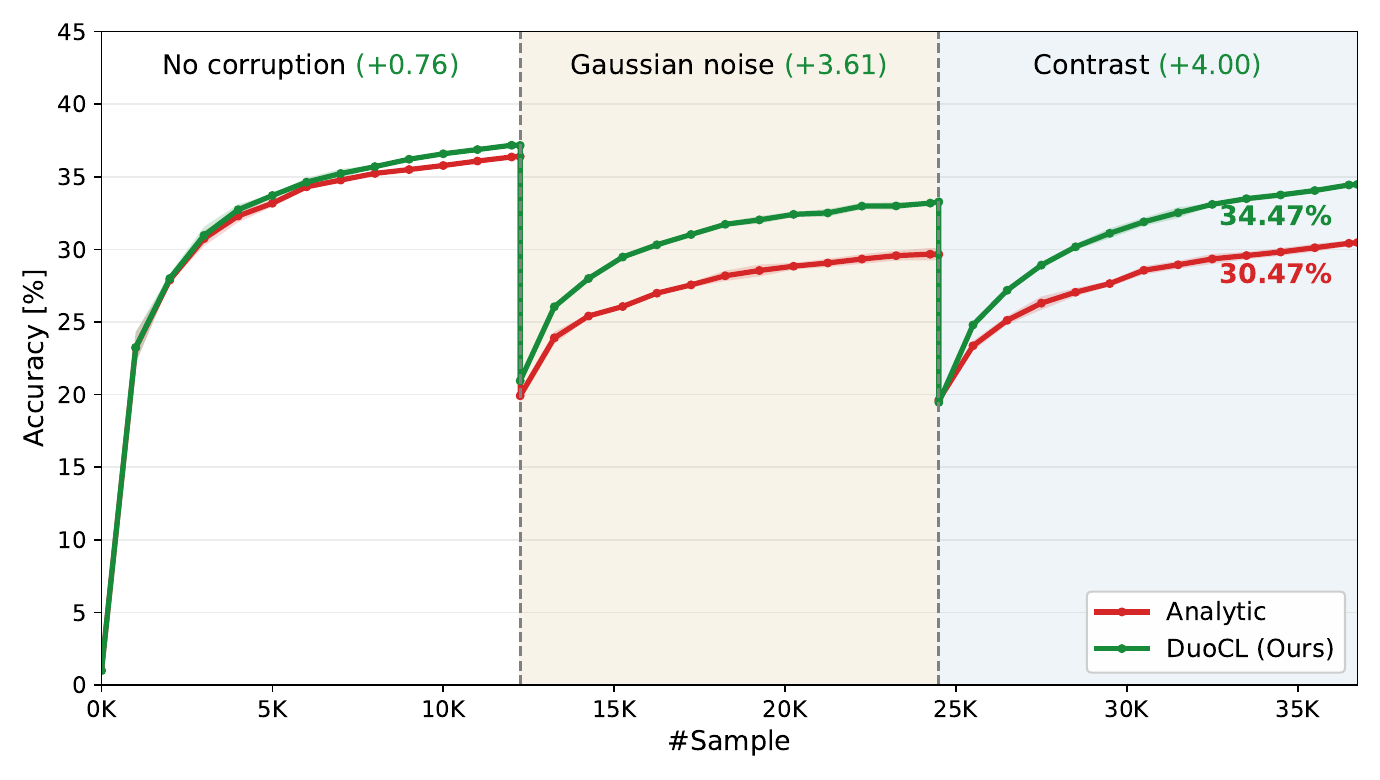}
    \caption{Accuracy under distribution shifts on CIFAR-C.} 
    \label{fig:cifar_c_shift}
\end{figure}

\begin{figure}
    \centering
    \includegraphics[width=0.9\columnwidth]{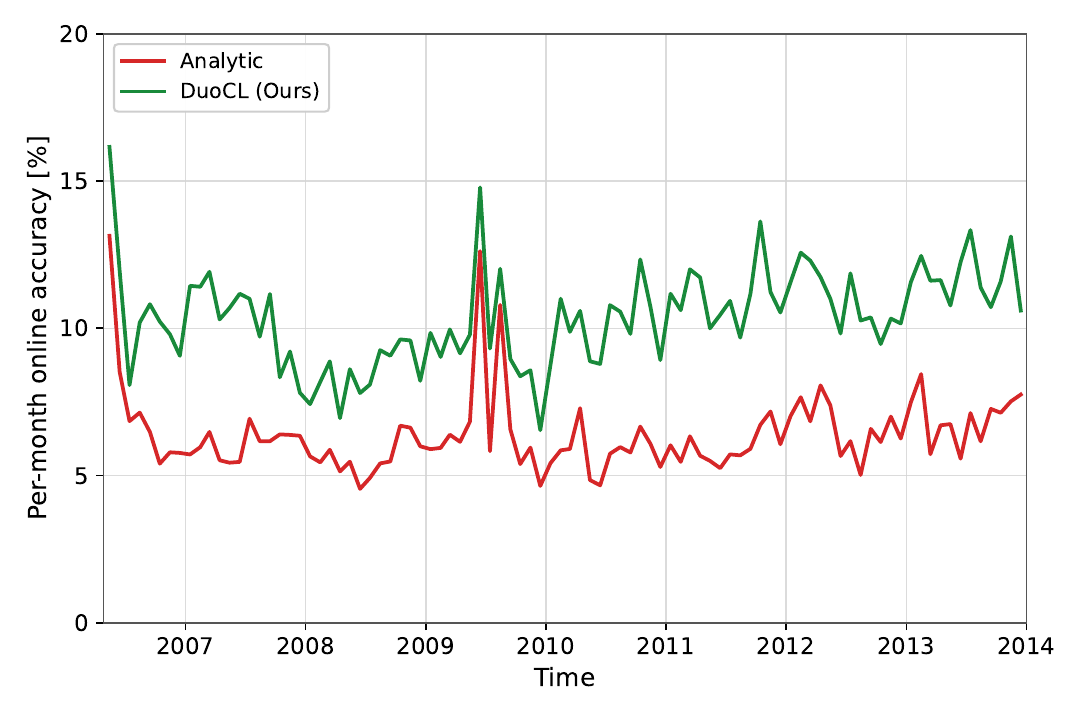}
    \caption{Comparison of online accuracy between DuoCL and Analytic on the CLOC dataset under naturally occurring temporal distribution shifts.}
    \label{fig:cloc_online_accuracy}
\end{figure}

To demonstrate the importance of dual-path learning under distribution shifts commonly encountered in OCL, we compare DuoCL with Analytic, a variant that uses only the fast path without backbone adaptation. We first evaluate both methods using ResNet-18 on a CIFAR-C stream consisting of clean images followed by Gaussian noise and contrast corruptions at severity level 4. On CIFAR-C, DuoCL improves test accuracy over Analytic by 3.61 and 4.00 percentage points under Gaussian noise and contrast corruption, respectively (Fig.~\ref{fig:cifar_c_shift}). On CLOC, using ResNet-50, DuoCL consistently achieves higher monthly online accuracy (Fig.~\ref{fig:cloc_online_accuracy}), supporting the benefit of backbone adaptation under naturally evolving distributions.


\subsubsection{Performance under Different OCL Settings}
\label{sec:ocl_settings}

\begin{table*}[t]
\centering
\caption{Performance comparison under Si-Blurry CIL
($\%$, $\uparrow$). Each dataset reports area under the
accuracy curve ($A_{\mathrm{AUC}}$) and final accuracy
($A_{\mathrm{Final}}$). Best and second-best non-UB
results are shown in \textbf{bold} and
\underline{underlined}, respectively.}
\label{tab:blurry_results}
\footnotesize
\setlength{\tabcolsep}{2.5pt}
\renewcommand{\arraystretch}{1.0}
\begin{tabular*}{\textwidth}{
@{\extracolsep{\fill}}llcccccc@{}}
\toprule
Backbone & Method
& \multicolumn{2}{c}{CIFAR-100}
& \multicolumn{2}{c}{ImageNet-R}
& \multicolumn{2}{c}{ImageNet-S} \\
\cmidrule(lr){3-4}
\cmidrule(lr){5-6}
\cmidrule(l){7-8}
& & $A_{\mathrm{AUC}}$ & $A_{\mathrm{Final}}$
  & $A_{\mathrm{AUC}}$ & $A_{\mathrm{Final}}$
  & $A_{\mathrm{AUC}}$ & $A_{\mathrm{Final}}$ \\
\midrule
\multirow{6}{*}{ViT-B/16}
& L2P~\cite{L2P}
& $43.01\pm9.37$ & $39.86\pm2.28$
& $22.71\pm1.86$ & $27.08\pm2.49$
& $10.02\pm0.42$ & $13.58\pm4.04$ \\
& MVP~\cite{MVP}
& $47.52\pm9.74$ & $44.49\pm0.93$
& $27.79\pm0.62$ & $31.64\pm1.77$
& $10.68\pm0.45$ & $13.99\pm1.73$ \\
& Online-LoRA~\cite{Online-LoRA}
& $60.12\pm5.79$ & $61.70\pm6.29$
& $45.05\pm1.59$ & $48.00\pm6.01$
& $30.81\pm2.09$ & $30.22\pm4.36$ \\
\cmidrule(l){2-8}
& Analytic
& $\underline{69.86}\pm6.03$
& $\underline{81.70}\pm0.00$
& $\underline{59.28}\pm0.31$
& $\underline{58.82}\pm0.07$
& $\underline{47.20}\pm2.50$
& $\underline{44.85}\pm0.24$ \\
& DuoCL (Ours)
& $\mathbf{74.74}\pm5.29$
& $\mathbf{86.63}\pm0.16$
& $\mathbf{60.95}\pm0.29$
& $\mathbf{61.65}\pm0.39$
& $\mathbf{48.53}\pm2.70$
& $\mathbf{46.15}\pm0.71$ \\
\cmidrule(l){2-8}
& \textit{Offline UB}
& \multicolumn{2}{c}{$89.50\pm0.04$}
& \multicolumn{2}{c}{$76.78\pm0.44$}
& \multicolumn{2}{c}{$63.82\pm0.02$} \\
\midrule
\multirow{6}{*}{ViT-S/16}
& L2P~\cite{L2P}
& $37.82\pm12.19$ & $30.88\pm1.39$
& $24.31\pm1.83$ & $21.83\pm2.13$
& $2.00\pm0.12$ & $3.61\pm1.08$ \\
& MVP~\cite{MVP}
& $40.31\pm9.52$ & $35.55\pm2.11$
& $27.04\pm1.09$ & $26.67\pm3.70$
& $2.27\pm0.14$ & $3.72\pm0.77$ \\
& Online-LoRA~\cite{Online-LoRA}
& $52.84\pm7.97$ & $58.72\pm1.44$
& $39.47\pm1.93$ & $36.61\pm4.63$
& $15.35\pm0.92$ & $20.18\pm1.84$ \\
\cmidrule(l){2-8}
& Analytic
& $\underline{64.70}\pm7.51$
& $\underline{77.93}\pm0.00$
& $\underline{44.85}\pm0.68$
& $\underline{43.56}\pm0.24$
& $\underline{33.35}\pm2.17$
& $\underline{32.06}\pm0.31$ \\
& DuoCL (Ours)
& $\mathbf{68.75}\pm7.24$
& $\mathbf{83.62}\pm0.35$
& $\mathbf{47.69}\pm0.62$
& $\mathbf{48.90}\pm0.79$
& $\mathbf{36.49}\pm2.40$
& $\mathbf{35.88}\pm0.78$ \\
\cmidrule(l){2-8}
& \textit{UB}
& \multicolumn{2}{c}{$86.55\pm0.01$}
& \multicolumn{2}{c}{$69.94\pm0.34$}
& \multicolumn{2}{c}{$59.28\pm0.11$} \\
\bottomrule
\end{tabular*}
\end{table*}

\begin{table}[t]
\centering
\caption{Final accuracy ($A_{\mathrm{Final}}$, \%, $\uparrow$)
under disjoint CIL. Best and second-best non-UB results
are shown in \textbf{bold} and \underline{underlined},
respectively.}
\label{tab:cil_results}
\scriptsize
\setlength{\tabcolsep}{2pt}
\renewcommand{\arraystretch}{1.10}
\resizebox{\columnwidth}{!}{%
\begin{tabular}{@{}llcccc@{}}
\toprule
Backbone & Method & CIFAR-100 & ImageNet-R
& ImageNet-S & CUB-200 \\
\midrule
\multirow{15}{*}{ViT-B/16}
& AGEM~\cite{agem}
& $12.67\pm1.87$ & $5.60\pm2.74$
& $0.16\pm0.04$ & $10.84\pm1.57$ \\
& ER~\cite{chaudhry2019continual}
& $44.85\pm1.83$ & $40.99\pm3.96$
& $30.21\pm0.70$ & $31.66\pm0.83$ \\
& EWC++~\cite{Rwalk}
& $10.61\pm0.74$ & $3.86\pm2.02$
& $0.32\pm0.28$ & $26.14\pm3.46$ \\
& MIR~\cite{MIR}
& $48.36\pm3.11$ & $41.51\pm2.99$
& $30.33\pm3.81$ & $31.64\pm2.97$ \\
& GDumb~\cite{GDumb}
& $41.00\pm19.97$ & $8.87\pm1.36$
& $1.65\pm0.22$ & $9.09\pm1.03$ \\
& PCR~\cite{PCR}
& $48.48\pm0.15$ & $46.11\pm3.03$
& $38.75\pm0.22$ & $41.11\pm1.43$ \\
& DER++~\cite{DER}
& $36.64\pm6.11$ & $30.90\pm8.04$
& $6.47\pm0.06$ & $26.61\pm1.27$ \\
& LODE (DER++)~\cite{LODE}
& $44.29\pm1.48$ & $42.20\pm6.46$
& $9.97\pm2.29$ & $39.20\pm4.25$ \\
& EMA (DER++)~\cite{EMA}
& $42.28\pm4.36$ & $41.75\pm1.98$
& $16.88\pm2.23$ & $35.26\pm3.31$ \\
& EMA (RAR)~\cite{EMA}
& $47.10\pm0.82$ & $30.04\pm0.33$
& $14.06\pm0.37$ & $33.34\pm1.11$ \\
& Online-LoRA~\cite{Online-LoRA}
& $49.40\pm1.36$ & $48.18\pm0.63$
& $47.06\pm0.24$ & $41.46\pm0.31$ \\
& FGH~\cite{michel2026online}
& $76.94\pm0.89$ & $64.85\pm0.71$
& $55.71\pm2.96$ & $75.22\pm1.55$ \\
& Analytic
& $\underline{85.55}\pm0.18$
& $\underline{68.24}\pm0.12$
& $\mathbf{59.12}\pm0.30$
& $\underline{82.86}\pm0.56$ \\
& DuoCL (Ours)
& $\mathbf{87.78}\pm0.16$
& $\mathbf{71.31}\pm0.14$
& $\underline{57.98}\pm0.49$
& $\mathbf{82.91}\pm0.54$ \\
& UB
& $89.50\pm0.04$ & $76.78\pm0.44$
& $63.82\pm0.02$ & $82.81\pm1.07$ \\
\midrule
\multirow{15}{*}{ViT-S/16}
& AGEM~\cite{agem}
& $7.43\pm2.15$ & $2.35\pm0.87$
& $2.75\pm2.86$ & $1.40\pm0.17$ \\
& ER~\cite{chaudhry2019continual}
& $31.91\pm2.06$ & $32.73\pm0.20$
& $19.53\pm1.44$ & $21.81\pm3.02$ \\
& EWC++~\cite{Rwalk}
& $6.80\pm2.13$ & $1.32\pm0.83$
& $4.08\pm3.24$ & $2.07\pm0.54$ \\
& MIR~\cite{MIR}
& $29.08\pm1.14$ & $34.73\pm0.29$
& $13.96\pm1.95$ & $22.95\pm1.12$ \\
& GDumb~\cite{GDumb}
& $10.87\pm4.94$ & $5.33\pm1.09$
& $2.09\pm0.32$ & $3.28\pm0.99$ \\
& PCR~\cite{PCR}
& $32.89\pm1.47$ & $21.96\pm0.27$
& $14.37\pm0.95$ & $22.28\pm2.73$ \\
& DER++~\cite{DER}
& $17.67\pm4.04$ & $22.17\pm4.27$
& $18.15\pm0.66$ & $29.53\pm2.37$ \\
& LODE (DER++)~\cite{LODE}
& $28.65\pm3.06$ & $31.65\pm0.72$
& $17.59\pm0.84$ & $26.81\pm0.89$ \\
& EMA (DER++)~\cite{EMA}
& $12.76\pm0.65$ & $20.89\pm3.05$
& $12.93\pm0.13$ & $35.79\pm5.27$ \\
& EMA (RAR)~\cite{EMA}
& $19.21\pm2.16$ & $16.11\pm0.35$
& $14.50\pm2.71$ & $34.53\pm1.04$ \\
& Online-LoRA~\cite{Online-LoRA}
& $32.16\pm0.24$ & $33.21\pm0.50$
& $22.45\pm0.43$ & $37.41\pm0.16$ \\
& FGH~\cite{michel2026online}
& $74.01\pm1.22$ & $52.39\pm4.37$
& $\underline{48.09}\pm2.28$ & $76.46\pm0.65$ \\
& Analytic
& $\underline{80.08}\pm0.06$
& $\underline{54.70}\pm0.07$
& $46.89\pm0.21$
& $\underline{83.22}\pm0.67$ \\
& DuoCL (Ours)
& $\mathbf{82.62}\pm0.11$
& $\mathbf{57.78}\pm0.01$
& $\mathbf{49.93}\pm0.18$
& $\mathbf{83.32}\pm0.56$ \\
& UB
& $86.55\pm0.01$ & $69.94\pm0.34$
& $59.28\pm0.11$ & $73.91\pm1.16$ \\
\bottomrule
\end{tabular}%
}
\end{table}

\begin{table}[t]
\centering
\caption{Final accuracy ($A_{\mathrm{Final}}$, $\%$, $\uparrow$)
under domain-incremental learning on CORe50~\cite{pmlr-v78-lomonaco17a}.
Best and second-best non-UB results are shown in
\textbf{bold} and \underline{underlined}, respectively.}
\label{tab:domain_results}
\footnotesize
\setlength{\tabcolsep}{5pt}
\renewcommand{\arraystretch}{1.05}
\begin{tabular}{@{}lcc@{}}
\toprule
Method & ViT-B/16 & ViT-S/16 \\
\midrule
AGEM~\cite{agem} & $80.15\pm2.97$ & $78.22\pm3.51$ \\
ER~\cite{chaudhry2019continual} & $85.85\pm1.35$ & $78.99\pm3.85$ \\
EWC++~\cite{Rwalk} & $78.65\pm6.51$ & $79.03\pm4.54$ \\
MIR~\cite{MIR} & $74.35\pm4.07$ & $86.49\pm0.81$ \\
GDumb~\cite{GDumb} & $77.20\pm3.49$ & $75.64\pm2.92$ \\
PCR~\cite{PCR} & $87.16\pm0.73$ & $75.20\pm1.48$ \\
DER++~\cite{DER} & $81.88\pm7.06$ & $89.33\pm0.62$ \\
LODE (DER++)~\cite{LODE} & $77.02\pm2.22$ & $83.48\pm5.84$ \\
L2P~\cite{L2P} & $87.97\pm0.37$ & $86.47\pm0.23$ \\
MVP~\cite{MVP} & $84.82\pm0.54$ & $79.85\pm0.33$ \\
Online-LoRA~\cite{Online-LoRA} & $\underline{93.71}\pm0.01$ & $\underline{90.96}\pm0.02$ \\
\midrule
Analytic & $93.57\pm0.00$ & $89.19\pm0.00$ \\
DuoCL (Ours) & $\mathbf{95.52}\pm0.08$ & $\mathbf{92.54}\pm0.14$ \\
\midrule
UB & $95.60\pm0.01$ & $93.56\pm0.01$ \\
\bottomrule
\end{tabular}
\end{table}

We evaluate DuoCL under disjoint class-incremental learning (CIL), where new classes are introduced sequentially across tasks and do not reappear in subsequent tasks. We consider CIFAR-100, ImageNet-R, ImageNet-S, and CUB-200 with the CIL setup using ViT-B/16 and ViT-S/16 backbones. We set the memory budget to $M=500$ samples and the available computational budget $F_{\mathrm{avail}}$ to the FLOPs corresponding to one training pass per incoming minibatch, consistent with prior work~\cite{Online-LoRA}. The first three datasets are divided into 10 tasks, while CUB-200 uses 5 tasks. UB represents the offline upper bound obtained through supervised fine-tuning on the complete dataset with i.i.d. samples. DuoCL achieves the highest final accuracy among OCL methods in seven of eight comparisons, improving upon Analytic by up to 3.08 percentage points (Table~\ref{tab:cil_results}).
Under domain-incremental learning (DIL), the input distribution changes while the class set remains fixed. On CORe50, DuoCL exceeds Online-LoRA~\cite{Online-LoRA} by 1.81 and 1.58 percentage points with ViT-B/16 and ViT-S/16, respectively (Table~\ref{tab:domain_results}).

We further evaluate DuoCL in the stochastic incremental-blurry (Si-Blurry) setting~\cite{MVP}, where class distributions change stochastically, and classes overlap across tasks. We consider CIFAR-100, ImageNet-R, and ImageNet-S, each divided into five tasks, using the same computational and memory budgets. DuoCL outperforms Analytic across all 12 $A_{\mathrm{AUC}}$ and $A_{\mathrm{Final}}$ comparisons (Table~\ref{tab:blurry_results}). On CIFAR-100, final accuracy improves by 4.93 and 5.69 percentage points with ViT-B/16 and ViT-S/16, respectively.


\subsubsection{Ablation on Individual Components}
\label{sec:component_ablation}

\begin{figure}[t]
    \centering
    \includegraphics[width=0.8\columnwidth,trim=0 21bp 0 0,clip]{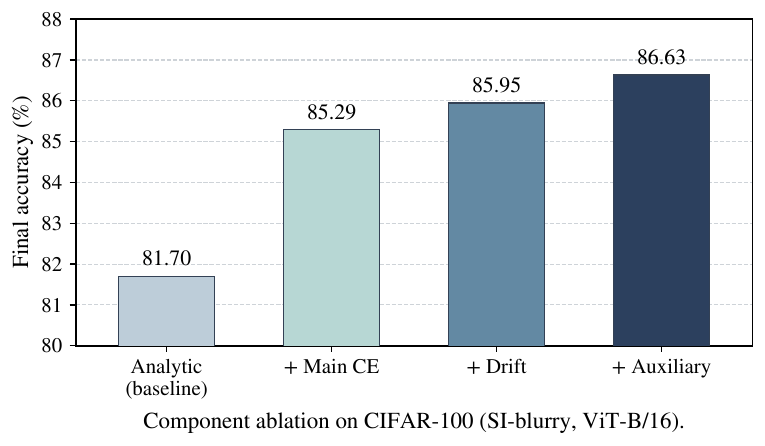}
    \caption{Component ablation on Si-Blurry CIFAR-100.}
    \label{fig:component_ablation}
\end{figure}

Figure~\ref{fig:component_ablation} presents the component ablation on CIFAR-100 under Si-Blurry using ViT-B/16. The fast-path-only baseline achieves 81.70\% final accuracy, increasing to 85.29\% with slow-path adaptation using the main cross-entropy loss, 85.95\% with drift regularization, and 86.63\% with the auxiliary loss, demonstrating the contribution of each component.

\section{Conclusion}
We presented DuoCL, a dual-path OCL framework that combines continuous analytic learning with periodic backbone adaptation under computational and memory constraints. DuoCL improves average online accuracy by 9.67 percentage points over existing baselines and demonstrates the benefits of backbone adaptation under distribution shifts and diverse OCL settings.

\section*{LLM Usage Statement}
OpenAI's ChatGPT and Anthropic's Claude were used to assist with editing throughout the manuscript, including grammar and clarity, as well as figure refinement, code implementation, and managing experiments. The authors reviewed and verified the final content and take full responsibility for the submitted work.
{
    \small
    \bibliographystyle{ieeenat_fullname}
    \bibliography{main}
}


\end{document}